\documentclass[11pt,twoside,twocolumn,a4paper]{article}

\usepackage{cvww}
\usepackage{times}
\usepackage{epsfig}
\usepackage{graphicx}
\usepackage{amsmath}
\usepackage{amssymb}
\usepackage{enumitem}
\usepackage{todonotes}
\usepackage{ifthen}
\usepackage{caption}
\usepackage{subcaption}

\usetikzlibrary{3d}

\usepackage{colortbl}
\usepackage{booktabs}

\newcommand{\deltaavg}{$<$$\delta^x_{avg}$}

\newcommand{\flownet}{\phi}
\newcommand{\flownetm}[1]{\flownet_{\text{#1}}}

\newcommand{\optflow}{\mathcal{F}}
\newcommand{\optflowp}[3]{\optflow_{#1}^{(#2,\, #3)}}

\newcommand{\certainty}{\rho}
\newcommand{\certaintyp}[3]{\certainty_{#1}^{(#2,\, #3)}}

\newcommand{\imdomain}{\Omega}

\newcommand{\flowvar}{\sigma}

\newcommand{\flowvarp}[3]{\flowvar_{#1}^{(#2,\, #3)}}
\newcommand{\flowvarpp}[4]{\flowvar_{#1}^{(#2,\, #3,\, #4)}}

\newcommand{\occl}{o}
\newcommand{\occlm}[1]{\occl_{\text{#1}}}
\newcommand{\occlp}[3]{\occlm{#1}^{(#2,\, #3)}}
\newcommand{\occlpp}[4]{\occlm{#1}^{(#2,\, #3, \, #4)}}

\newcommand{\point}{\boldsymbol{p}}
\newcommand{\pointp}[1]{\point_{#1}}

\newcommand{\score}{s}
\newcommand{\scorep}[2]{\score^{(#1,\, #2)}}

\newcommand{\flowseq}{\mathcal{S}}
\newcommand{\predset}{\mathcal{P}}

\newcommand{\dist}{\Delta}
\newcommand{\distp}[1]{\dist_{#1}}

\newcommand{\jdma}{i_{M}}
\newcommand{\interf}{i}

\newcommand{\thresh}{\theta}

\usepackage[pagebackref=true,breaklinks=true,bookmarks=false]{hyperref}
\graphicspath{{figures/}}

\cvwwfinalcopy 

\def\cvwwPaperID{14} 

\ifcvwwfinal\fi

\begin{document}

\title{Dense Matchers for Dense Tracking}

\author{Tomáš Jelínek, Jonáš Šerých, Jiří Matas\\
CMP Visual Recognition Group, Faculty of Electrical Engineering,\\Czech Technical University in Prague\\
{\tt\small \{tomas.jelinek,serycjon,matas\}@fel.cvut.cz}
}

\maketitle
\ifcvwwfinal\thispagestyle{fancy}\fi

\begin{abstract}
      Optical flow is a useful input for various applications, including 3D reconstruction, pose estimation, tracking, and structure-from-motion. Despite its utility, the problem of dense long-term tracking, especially over wide baselines, has not been extensively explored. This paper extends the concept of combining multiple optical flows over logarithmically spaced intervals as proposed by MFT. We demonstrate the compatibility of MFT with two dense matchers, DKM and RoMa. Their incorporation into the MFT framework optical flow networks yields results that surpass their individual performance. Moreover, we present simple yet effective ensembling strategies that prove to be competitive with more sophisticated, non-causal methods in terms of position prediction accuracy, highlighting the potential of MFT in long-term tracking applications.
\end{abstract}


\section{Introduction}

Obtaining point-to-point correspondences is a classical task in computer vision, useful for a wide range of applications including tracking, structure-from-motion, and localization. Despite the extensive research in wide baseline stereo methods, including those with a time baseline, the domain of dense point correspondences in videos has not been explored until recently \cite{neoral2024mft, wang2023omnimotion}.
The emergence of the TAP-Vid dataset~\cite{doersch2022tap} has further fueled interest in long-term point-tracking methods.

Point-trackers usually~\cite{doersch2023tapir, karaev2023cotracker, sand2008particle, doersch2022tap} track sparse sets of points. However, dense correspondences are useful in various applications, such as video editing, object tracking, and 3D reconstruction. While optical flow techniques provide dense correspondences, they are typically limited to pairs of consecutive frames. 

Long-term dense tracking has been recently addressed by Neoral~\etal~\cite{neoral2024mft} MFT tracker, which computes optical flow not only for consecutive frames but also for pairs of more temporally distant frames, including flow computation between the reference and every other frame of the video.
At every frame, optical flow is computed w.r.t. the previous, first, and a constant number of logarithmically spaced frames. Such approach is linear in the number of frames and thus not computationally prohibitive.

In the original MFT\cite{neoral2024mft}, all optic flow computations are based on RAFT \cite{teed2020raft}, which has performed  well in both standard benchmarks \cite{butler2012sintel, menze2018kitti} and in applications.
However, the RAFT optical flow network was trained on pairs of consecutive frames, which is likely sub-optimal for large baselines. 

Recently, dense matchers such as DKM~\cite{edstedt2023dkm} and RoMa~\cite{edstedt2023roma} have been published. This development opens the possibility to apply the MFT framework with different dense matchers, or to use RAFT for pairs of frames with short temporal, and thus probably spatial, baseline. The only requirement of the MFT ``meta optic flow algorithm'' is that the basis dense two view optic flow or matcher provides confidence in its predictions.


In this paper, we evaluate the MFT approach with the DKM and RoMa matchers instead of RAFT. We show that both of these matchers  provide accurate matches, but inaccurate occlusion predictions. Addressing the strengths and weaknesses of optical-flow-based and dense-matching-based methods, we propose a combined tracker, that outperforms the original MFT design.


In summary, our contributions are:
(1)~We show how to adapt dense matchers DKM and RoMa for use in the MFT framework, and experimentally evaluate their performance.
(2)~We show that the MFT algorithm outperforms both direct flow between the first and the current frame, and the chaining of optical flows computed on consecutive frames for RAFT, DKM, and RoMa.
(3)~Based on better results of RoMa over DKM in our experiments, we propose a dense long-term tracker that combines the strengths of RAFT-based MFT and RoMa-based MFT.

\section{Related Work}
\label{sec:sota}

\paragraph{Tracking, 3D Reconstruction, and SLAM}\hspace{-0.65em}
Object tracking algorithms~\cite{bolme2010visual, kalal2011tracking, danelljan2018atom} traditionally outputted the track of an object specified in the first frame in the form of bounding boxes. Later, the focus shifted towards segmentation-based tracking~\cite{kristan2020eighth, perazzi2016davis, lukezic2019d3s}.

Modern model-free trackers based on differentiable rendering~\cite{wen2023bundlesdf, rozumnyi2023tracking}, that can simultaneously track and reconstruct any object specified in the first frame are naturally able to provide point-to-point correspondences for the tracked object; however, to the best of our knowledge, they can track a single object only or require multi-camera input \cite{luiten2023dynamic}. Additionally, recent methods ~\cite{yang2021lasr, yang2022banmo, wu2023dove}, involving differentiable rendering of neural radiance fields (NeRFs) ~\cite{mildenhall2021nerf}, show potential in creating deformable 3D models for point tracking. Nonetheless, the extensive computational demands of these methods limit their practical applicability in real-world scenarios.

The traditional SLAM methods~\cite{schoenberger2016sfm} produced sparse point clouds. Later on,  semi-dense~\cite{engel2013semi, teed2021droid} SLAM methods appeared. Some SLAM-based trackers,  ~\cite{gladkova2022directtracker} can densely estimate point positions in static scenes, and recent advances in differentiable rendering opened the avenue for differentiable-rendering-based monocular SLAMs~\cite{rosinol2023nerf} but their application remains constrained to static scenes.\vspace{-1em}

\paragraph{Optical Flow}\hspace{-0.65em} estimation is a classical problem in computer vision, with the early works~\cite{lucas1981iterative, horn1981flow} relying on the brightness-constancy assumption. With the advent of deep neural networks, the focus shifted towards learning-based approaches~\cite{dosovitskiy2015flownet, sun2018pwc, Dosovitskiy2017FlowNet2, teed2020raft, huang2022flowformer} trained on synthetic data.

Optical flow estimation in state-of-the-art methods, exemplified by RAFT~\cite{teed2020raft} and FlowFormer~\cite{huang2022flowformer}, is achieved through the analysis of a 4D correlation cost volume, considering features of all pixel-pairs. These techniques excel in densely estimating flow between consecutive frames, yet they encounter challenges in accurately determining flow across distant frames, particularly in scenarios with large displacements or significant object deformation.

Multi-step-flow algorithms~\cite{crivelli2012optical, crivelli2012multi, crivelli2014robust} address the limitations of concatenation-based approaches for long-term dense point tracking.  These algorithms create extended dense point tracks by merging optical flow estimates across variable time steps, effectively managing temporarily occluded points by bypassing them until their re-emergence. However, their dependence on the brightness constancy assumption renders them less effective over distant frames. Subsequent works in multi-step-flow, such as the multi-step integration and statistical selection (MISS) approach by Conze et al.~\cite{conze2014dense, conze2016multi}, further refine this process. This approach relies on generating a multitude of candidate motion paths from random reference frames, with the best path selected through a global spatial smoothness optimization process. However, this strategy makes these methods computationally demanding.
Although certain optical flow techniques~\cite{ilg2018occlusions, neoral2018continual, hur2019iterative, zhao2020maskflownet, liu2021oiflow, zhang2022parallel} address occlusions and flow uncertainty, most leading optical flow methods, influenced by standard benchmarks like those in Butler~\etal~\cite{butler2012naturalistic} and Menze~\etal~\cite{menze2015object}, do not detect occlusions. Jiang~\etal~\cite{jiang2021learning}, building on RAFT~\cite{teed2020raft}, has taken a different approach in which they handle occlusion implicitly by computing hidden motions of the occluded objects. However, the method still falls short in the context of tracking dynamic, complex motions.

We now describe in greater depth three methods that are most relevant to our paper: RAFT~\cite{neoral2024mft}, DKM~\cite{edstedt2023dkm}, and RoMa~\cite{edstedt2023roma}. While the latter two are in fact dense matchers, we will use the term interchangeably with long-ranged optical flow estimation with occlusion prediction.
\vspace{-1em}

\paragraph{MFT}\hspace{-0.65em} extends optical flow into dense long-term trajectories by constructing multiple chains of optical flows and selecting the most reliable one~\cite{neoral2024mft}.
The flow chains consist of optical flow computed both between consecutive frames, and between more distant frames, which allows for re-detecting points after occlusions.
The intervals between distant frames are chosen to be logarithmically spaced.

MFT extends the RAFT optical flow method with two heads, estimating occlusion and uncertainty for each flow vector.
Like the optical flow, the uncertainty and the occlusion are accumulated over each chain, and the non-occluded flow chain with the least overall uncertainty is selected as the most reliable candidate.
The long-term tracks of different points thus chain possibly different sequences of optical flows.
This strategy on one hand takes into account that changes in appearance and viewpoint gradually accumulate over time, which makes it more reliable to chain flows on easier-to-match frames rather than estimating matches directly between the template and the current frame.
On the other hand, short chains containing longer jumps with low uncertainty result in less error accumulation.

\vspace{-1em}

\paragraph{DKM}\hspace{-0.65em} proposed by Edstedt~\etal~\cite{edstedt2023dkm}, a dense point-matching method, employing a ResNet~\cite{he15resnet}-based encoder pre-trained on ImageNet-1K~\cite{russakovsky2015imagenet} for generating both fine and coarse features. The coarse features undergo sparse global matching, modeled as Gaussian process regression, to determine embedded target coordinates and certainty estimates. Fine features are refined using CNN refiners, following a methodology similar to Truong~\etal~\cite{truong2020glu} and Shen~\etal~\cite{shen2020ransac}. DKM's match certainty estimation relies on depth consistency, necessitating 3D supervision. The process concludes by filtering matches below a certainty threshold of $0.05$ weighted sampling for match selection. Edstedt~\etal~\cite{edstedt2023dkm} released outdoor and indoor models trained on MegaDepth~\cite{MegaDepthLi18}) and ScanNet~\cite{dai2017scannet} respectively.
\vspace{-1em}

\paragraph{RoMa}\hspace{-0.65em} similarly to DKM, RoMa~\cite{edstedt2023roma} is a dense matching method that provides pixel displacement vectors along with their estimated certainty, building upon the foundation set by DKM~\cite{edstedt2023dkm}. RoMa differentiates itself by employing a two-pronged approach for feature extraction: using frozen DINOv2~\cite{oquab2023dinov2} for sparse features and a specialized ConvNet with a VGG19 backbone~\cite{simonyan14vgg} for finer details. Unique to RoMa is their transformer-based match decoder, which matches features through a regression-by-classification approach, better handling the multimodal nature of coarse feature matching. In contrast to DKM, RoMa's pipeline omits the use of dense depth maps for match certainty supervision, relying instead on pixel displacements for match supervision. Their model is trained on datasets like MegaDepth~\cite{MegaDepthLi18} and ScanNet~\cite{dai2017scannet}, similar to DKM.
\vspace{-1em}

\paragraph{Long-Term Point Tracking}\hspace{-0.65em}aiming to track a set of physical points in a video has emerged significantly since the release of TAP-Vid~\cite{doersch2022tap}. The dataset's baseline method TAP-Net~\cite{doersch2022tap} computes a cost volume for each frame, employing a technique akin to RAFT's approach~\cite{teed2020raft}. It focuses on tracking individual query points. PIPs~\cite{harley2022particle} takes this approach to an extreme by completely trading off spatial awareness about other points for temporal awareness within fixed-sized temporal windows, making it unable to re-detect the target after longer occlusions. TAPIR~\cite{doersch2023tapir} combines TAP-Net's track initialization with PIPs' refinement while removing the PIPs' temporal chunking, using a time-wise convolution instead. CoTracker~\cite{karaev2023cotracker} models the temporal correlation of different points via a sliding-window transformer, modeling multiple tracks' interactions. While these methods are designed for sparse tracking, they can provide dense tracks by querying all points in the first frame.

Notably, differentiable rendering has been leveraged in recent approaches, with OmniMotion representing 3D points' motion implicitly using learned bijections~\cite{wang2023omnimotion} enabling it to provide dense tracks. Alternative methods like~\cite{luiten2023dynamic} which models the scene as temporally-parametrized Gaussians\cite{kerbl3Dgaussians}. However, these methods have their limitations, such as OmniMotion's quadratic complexity and the multi-camera requirement of~\cite{luiten2023dynamic}.

\section{Method}  

For a stream $\{\mathcal{I}_1,\, ...,\, \mathcal{I}_N \}$ of $N$ video frames defined on a common image domain $\imdomain$, we denote the optical flow between frames $i$ and $j$ as $\optflowp{}{i}{j}$.
Moreover, we use $\flowvarp{}{i}{j} \in \mathbb{R}_{+}^{\imdomain}$ to denote the estimated flow variance, and $\certaintyp{}{i}{j} \in [0,\, 1]^{\imdomain}$ to represent the estimated certainty of $\optflowp{}{i}{j}$.
Finally, occlusion score $\occlp{}{i}{j} \in [0,\, 1]^\Omega$ denotes the estimated probability of pixels appearing in frame $i$ being occluded in frame $j$.
To simplify notation, although $\optflowp{}{i}{j}$, $\certaintyp{}{i}{j}$, and $\flowvarp{}{i}{j}$ are 2D or 3D tensors, we will use these symbols to denote their values at a specific point $\point = (x,\,y)$ in the image. Moreover, for every point $\pointp{i}$ in frame~$i$, its predicted position $\boldsymbol{p}_j$ in frame $j$ relates to the optical flow $\optflowp{}{i}{j}$ as follows: 
\begin{equation}
    \boldsymbol{p}_{j} = \boldsymbol{p}_i + \optflowp{}{i}{j}(\boldsymbol{p}_i)\label{eq:pointflow}.
\end{equation}

Let us denote by $\flownetm{RAFT}$, $\flownetm{DKM}$,\, $\flownetm{RoMa}$,\, $\flownetm{MFT}$ the functions computed by RAFT, DKM, RoMa, and MFT respectively. By RAFT we mean the MFT's adaptation of RAFT with additional uncertainty and occlusion heads~\cite{neoral2024mft}. The output vectors of these methods are as follows: 
\begin{gather}
\flownetm{RAFT} = (\optflowp{}{i}{j},\, \flowvarp{}{i}{j},\, \certaintyp{}{i}{j})\\
\flownet_{W}=(\optflowp{}{i}{j},\, \certaintyp{}{i}{j})\\
\flownetm{MFT} = (\optflowp{}{i}{j},\, \flowvarp{}{i}{j},\, \occlp{}{i}{j}),
\end{gather}
where $W$ is one of the wide-baseline methods, either DKM or RoMa.


\subsection{MFT Flow Chaining}\label{subsec:chaining}

MFT~\cite{neoral2024mft} achieves long-term optical flow estimation by combining multiple optical flows. These flows are obtained from $\flownetm{RAFT}$ over logarithmically spaced distances. When estimating the flow $\optflowp{}{1}{j}$, MFT utilizes a sequence of intermediate flows. This sequence, denoted as $\flowseq$, comprises flows $\optflowp{}{j - \distp{1}}{j},\, ...,\, \optflowp{}{j - \distp{K}}{j}$. Here, $\distp{i}$ represents logarithmic spacing and is defined as $2^{i - 1}$ for $i < K$, with $\distp{K} = j-1$. We limit the number of intermediate flows, denoted by $K$, to a maximum of 5 and ensure that $j - \distp{K - 1} > 1$.

Additionally, MFT employs a scoring function for evaluating the quality of the intermediate flows chaining for each image point~$\pointp{1}$ in the reference frame $1$.
The scoring function $\scorep{j - \distp{k}}{j}$ utilizes chaining of estimated flow variances and occlusion scores over an intermediate frame $\interf$:
\begin{align}
 \flowvarpp{}{1}{\interf}{j}(\pointp{1}) &= \flowvarp{MFT}{1}{\interf}(\pointp{1}) + \flowvarp{}{\interf}{j} (\pointp{\interf}), \\
\occlpp{}{1}{\interf}{j}(\pointp{1}) &= \max \{  \occlp{MFT}{1}{\interf}(\pointp{1}),\, \occlp{}{\interf}{j}(\pointp{\interf})  \},
\end{align}
The point $\pointp{i}$ is computed using $\optflowp{MFT}{1}{i}$ and the relation in Equation \ref{eq:pointflow}. The scoring function is then defined as $\scorep{j - \distp{k}}{j}(\pointp{1})=-\flowvarpp{}{1}{j - \distp{k}}{j}(\pointp{1})$.
If the chained occlusion score $\occlpp{}{1}{j - \distp{k}}{j}(\pointp{1})$ exceeds an occlusion threshold $\thresh_{\occl}$, we set $\scorep{j - \distp{k}}{j}(\pointp{1}) = -\infty$.
This score is used to select the best flow for every point $\pointp{1}$, that is the flow with the lowest estimated variance computed on chains that do not contain occluded points. 

\begin{figure}[htbp]
\centering

\tikzset{fontscale/.style = {font=\relsize{#1}}
    }

\begin{tikzpicture}[
pnode/.style={circle, draw=black, thick, minimum size=15mm, font=\scriptsize, scale=0.8, inner sep=0},
arcstyle/.style={->, dashed, thick, font=\scriptsize}
]
    \edef \diff {1.1}
    \edef \dotc {0.1}

    \node at (0, 1 * \diff + \dotc) {\vdots};
    \node[pnode] (nodeorig) at (0,2 * \diff) {$\pointp{1}$};
    
    \node[pnode] (node1) at (0,0) {$\pointp{j - \Delta_{K-1}}$};
    
    \node at (0, -1 * \diff + \dotc) {\vdots};
    \node[pnode] (node3) at (0,-2 * \diff) {$\pointp{j - \Delta_1}$};
    
    \node at (0, -3 * \diff + \dotc) {\vdots};
    \node[pnode] (nodej) at (0,-4 * \diff) {$\pointp{j}$};

    \node[draw=black, font=\scriptsize, align=center] (nodesel) at (3,-4 * \diff) {
    \parbox{3.61cm}{  
        \scriptsize $\predset=\{j - \distp{k}|\ 1 \leq k \leq K\}$
        \vspace{5pt}  
        \\$i_M=\arg\max_{i \in \predset}\scorep{i}{j}(\pointp{1})$
    }
};

    \draw[arcstyle] (nodeorig) to [bend right=50] node[midway, left=0.25] {$\optflowp{MFT}{1}{j-\distp{K-1}}$} (node1);
    \draw[arcstyle] (nodeorig) to [bend right=50] node[midway, left=0.05] {$\optflowp{MFT}{1}{j-\distp{1}}$} (node3);

    \draw[arcstyle] (nodeorig) to [bend left=60] node[pos=0.035, yshift=7pt, right=0.45, align=left] {$\optflowp{}{j-\distp{K}}{j},$\\$\scorep{j-\distp{K}}{j}$} (nodesel);
    \draw[arcstyle] (node1) to [bend left=50] node[pos=0.05, yshift=10pt, right=0.00, align=left] {$\optflowp{}{j-\distp{K-1}}{j},$\\$\scorep{j-\distp{K-1}}{j}$} (nodesel);
    \draw[arcstyle] (node3) to [bend left=50] node[pos=0.25, above=0.12, align=left] {$\optflowp{}{j-\distp{1}}{j},$\\$\scorep{j-\distp{1}}{j}$} (nodesel);

    \draw[arcstyle] (nodesel) to [bend left=50] node[pos=0.15, below=0.12, align=left] {$\optflowp{}{i_M}{j}$} (nodej);

\end{tikzpicture}
\caption{Illustration of the MFT flow chaining as defined in Equation \ref{eq:flowchain}. The optical flows are evaluated on the points in the outbound nodes of their respective arcs.}
\label{fig:mftlogic}
\end{figure}
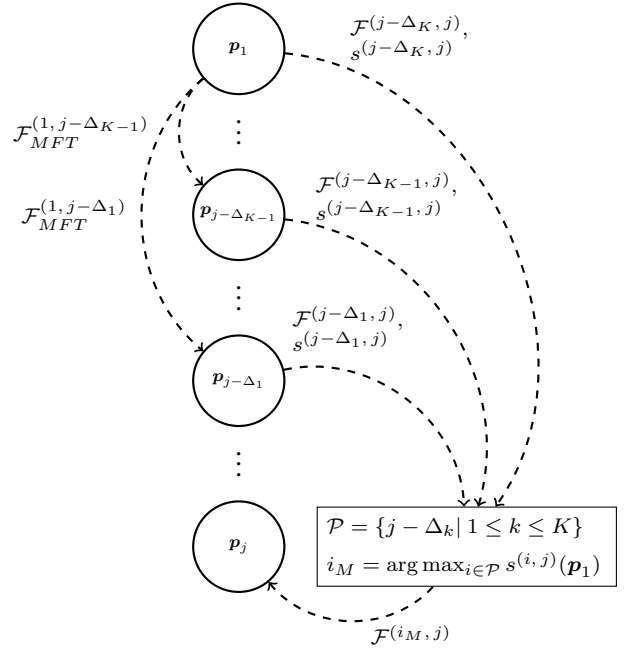

MFT computes long-term flow for any point $\boldsymbol{p}_1$ in the reference frame $1$ iteratively via chaining as
\begin{align}
    &\optflowp{MFT}{1}{j}(\pointp{1}) = \optflowp{MFT}{1}{i_M}(\pointp{1}) + \optflowp{}{\jdma}{j}(\pointp{\jdma}),\label{eq:flowchain}
\end{align}
where $\jdma \in \{j - \distp{k}\ |\ 1 \leq k \leq K\} $ such that the score $\scorep{\jdma}{j}(\pointp{1})$ is maximal. Again, the point $\pointp{\jdma}$ is obtained using $\optflowp{MFT}{1}{i_M}(\pointp{1})$ and Equation~\ref{eq:pointflow}.
$\optflowp{}{\jdma}{j}$ is the flow obtained from an arbitrary method that can also estimate its variance $\flowvarp{}{\jdma}{j}$ and occlusion score~$\occlp{}{\jdma}{j}$. The flow chaining is visualized in Figure \ref{fig:mftlogic}.

The estimated variance and occlusion score for frame $j$ are then obtained from the chain over frame~$\jdma$ as 
$\flowvarp{MFT}{1}{j}(\pointp{1}) = \flowvarpp{}{1}{\jdma}{j}(\pointp{1})$, respectively
    $\occlp{MFT}{1}{j}(\pointp{1}) = \occlpp{}{1}{\jdma}{j}(\pointp{1})$.
A pixel observed in frame $i$ is considered occluded in frame $j$ if its value $\occlp{MFT}{i}{j}$ is above a threshold $\thresh_{\occlm{}}$. In practice, we set different thresholds for different backbone networks as we discuss in Subsection \ref{subseq:integration}.

\subsection{Integration of DKM and RoMa}\label{subseq:integration}

As we mentioned in the Introduction, we make the conjecture that training RAFT  for optical flow prediction on consecutive video frames is suboptimal for wide baselines.
We therefore integrate DKM and RoMa, capable of handling wider baselines. However, integrating these methods with MFT poses certain challenges due to their incompatible outputs.

In the first place, neither RoMa nor DKM provides an occlusion score $\occl$, but only an estimate of the flow prediction certainty $\certainty$. We therefore artificially set their occlusion scores as $\occl = 1 - \certainty$. Furthermore, although $\flowvar$ and $\certainty$ both represent the quality of estimated optical flow, they are not directly comparable. But in order to integrate them into the MFT framework, we need to converse between them.


Through empirical analysis, we established a flow certainty threshold $\thresh_\certainty$. When $\certainty$ exceeds this threshold, we deem the optical flow reliable, assigning $\flowvar=0$. Conversely, when $\certainty$ is below this threshold, $\flowvar$ is set to $1000$, correlating higher uncertainties with increased variances in predicted flow. Additionally, we observed that while $\occlm{MFT}$, $\occlm{DKM}$, and $\occlm{RoMa}$ fundamentally represent the same concept, their respective occlusion thresholds $\thresh_{\occlm{RAFT}}$ and $\thresh_{\occlm{RoMa}}$ vary. In our experiments in Section~\ref{sec:experiments}, we use
\begin{gather}
\thresh_{\occlm{RAFT}}=0.02,\ \thresh_{\occlm{DKM}}=\thresh_{\occlm{RoMa}}=0.95.
\end{gather}

\begin{figure*}[htbp]
\centering

\begin{minipage}{\textwidth}
  \begin{subfigure}{.45\textwidth}
    \centering
    \includegraphics[width=\linewidth]{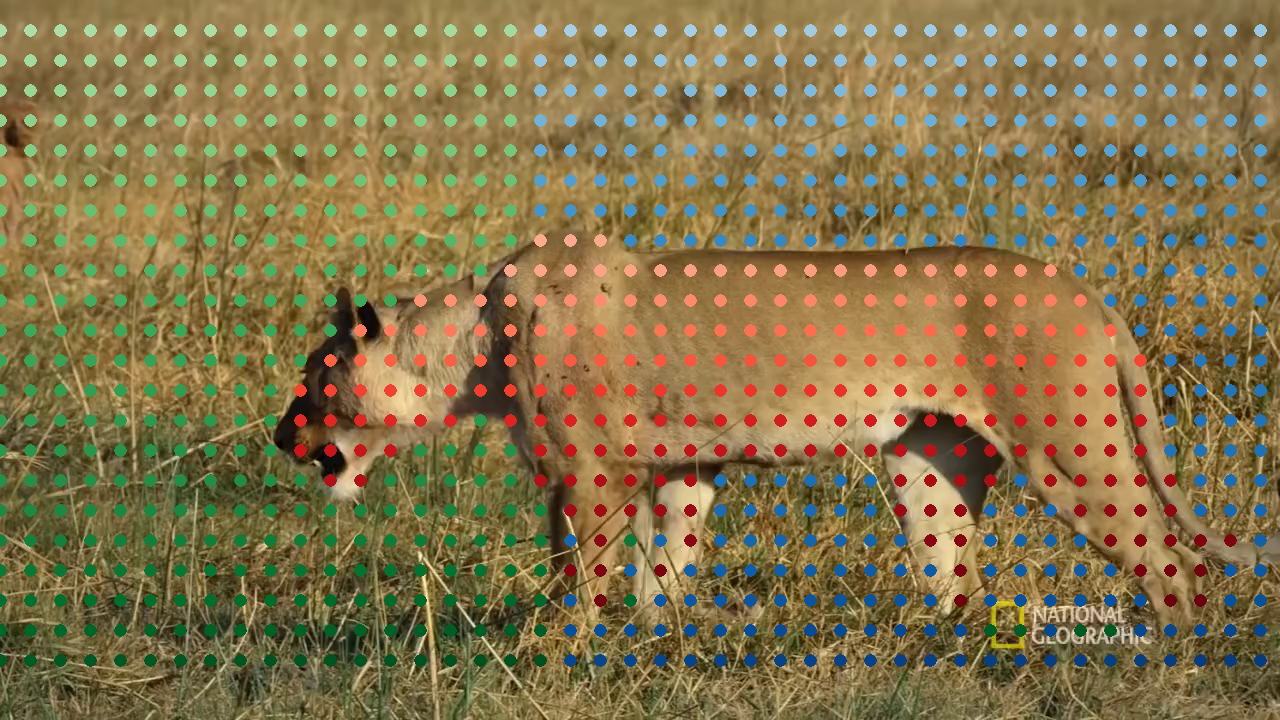}
    \caption{Reference frame}
  \end{subfigure}\hfill
  \begin{subfigure}{.45\textwidth}
    \centering
    \includegraphics[width=\linewidth]{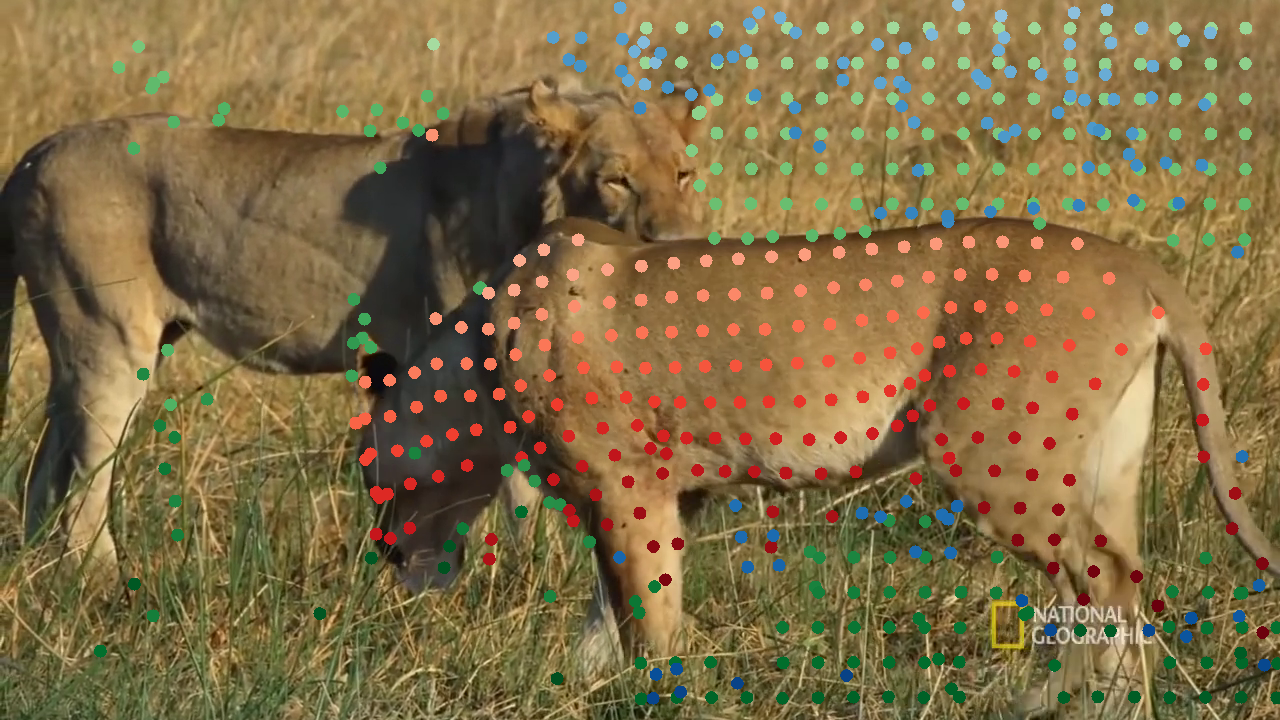}
    \caption{RAFT-based MFT Strategy.}
    \label{subfig:mftraft}
  \end{subfigure}
\end{minipage}

\smallskip

\begin{minipage}{\textwidth}
  \begin{subfigure}{.45\textwidth}
    \centering
    \includegraphics[width=\linewidth]{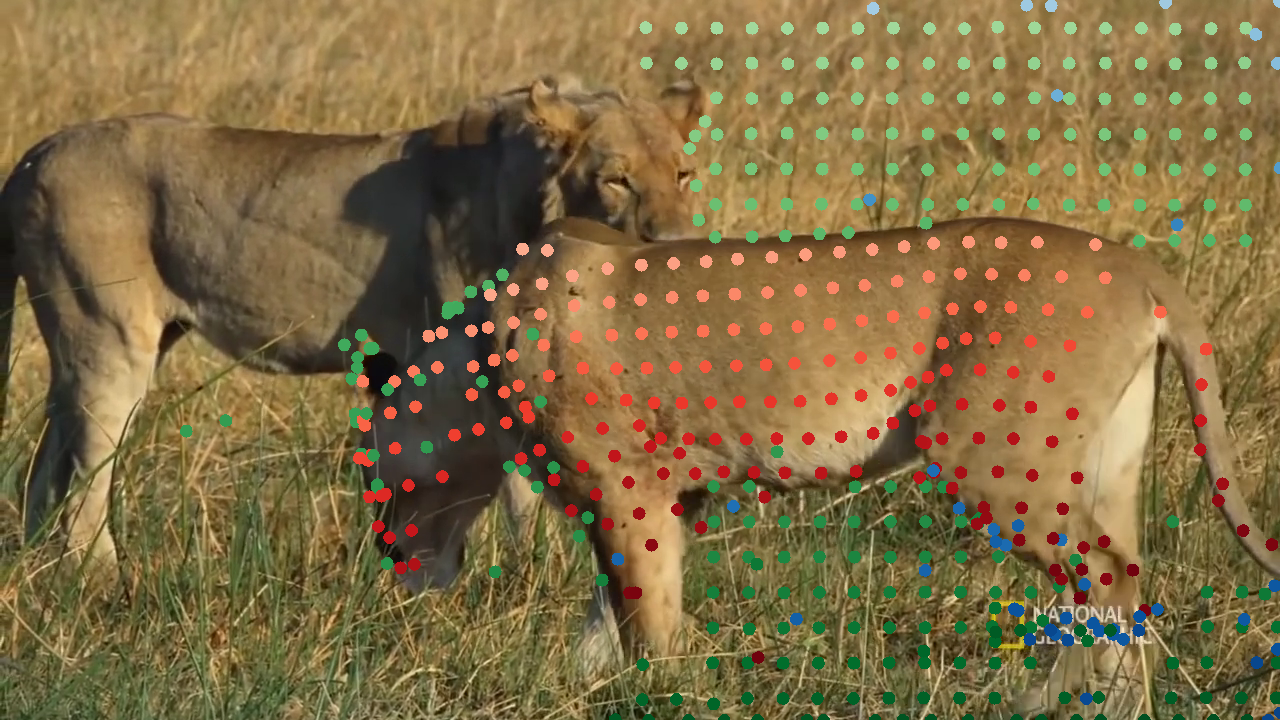}
    \caption{RoMa-based MFT Strategy.}
    \label{subfig:mftroma}
  \end{subfigure}\hfill
\begin{subfigure}{.45\textwidth}
    \centering
    \includegraphics[width=\linewidth]{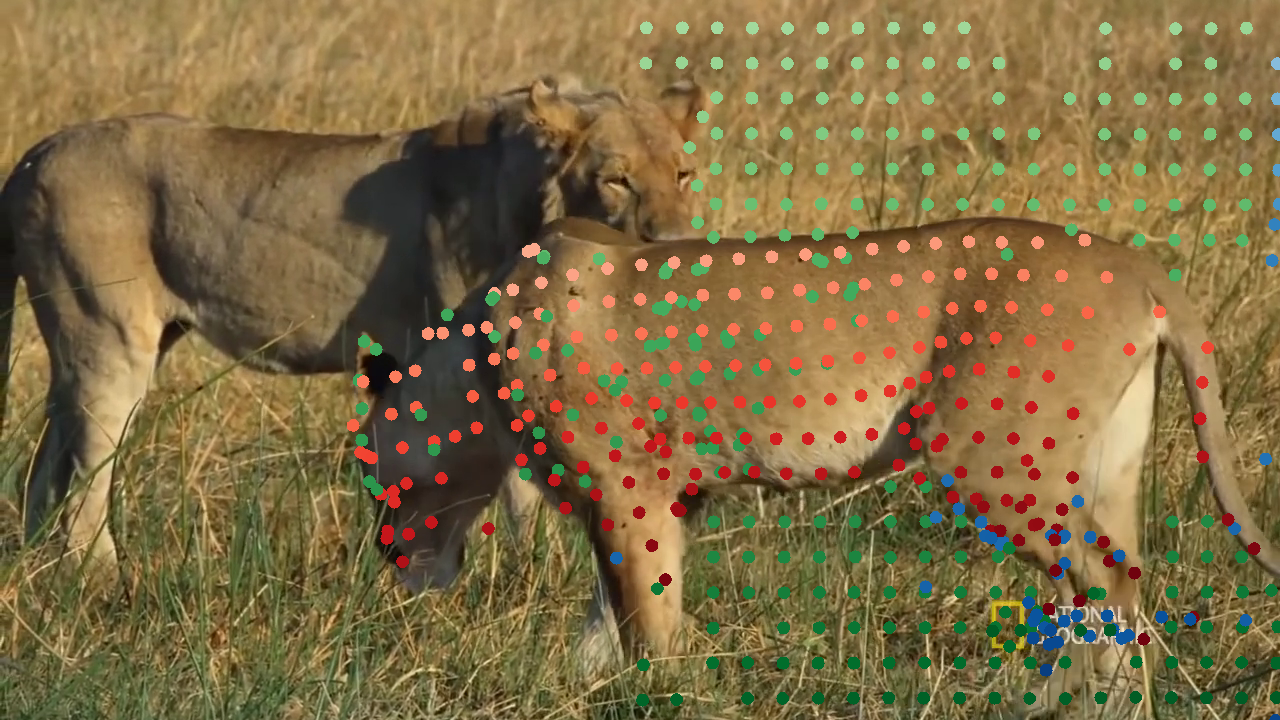}
    \caption{DKM-based MFT Strategy.}
    \label{subfig:mftdkm}
\end{subfigure}
\textbf{}
\end{minipage}

\smallskip

\begin{minipage}{\textwidth}
  \begin{subfigure}{.45\textwidth}
    \centering
    \includegraphics[width=\linewidth]{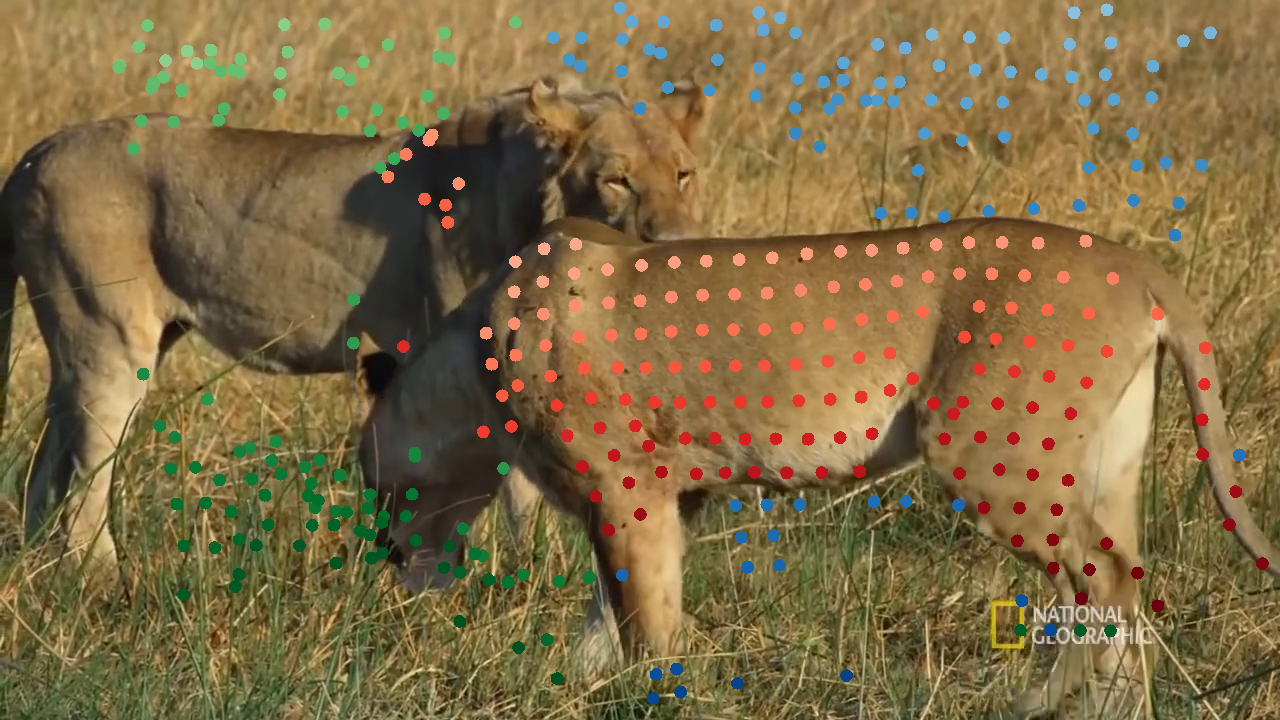}
    \caption{Direct matching between frames \#0 and \#140 using RAFT.}
    \label{subfig:raftdirect}
  \end{subfigure}\hfill
\begin{subfigure}{.45\textwidth}
    \centering
    \includegraphics[width=\linewidth]{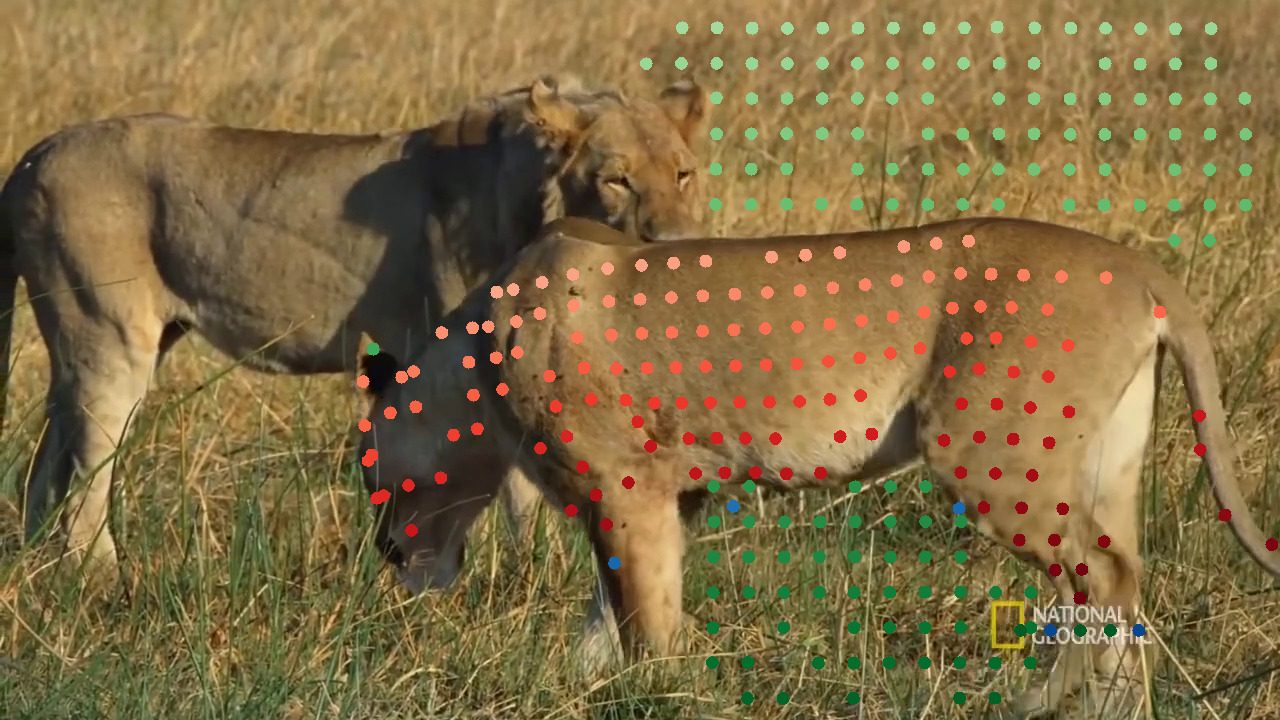}
    \caption{Direct matching between frames \#0 and \#140 using RoMa.}
    \label{subfig:romadirect}
  \end{subfigure}
\end{minipage}

\smallskip

\begin{minipage}{\textwidth}
  \begin{subfigure}{.45\textwidth}
    \includegraphics[width=\linewidth]{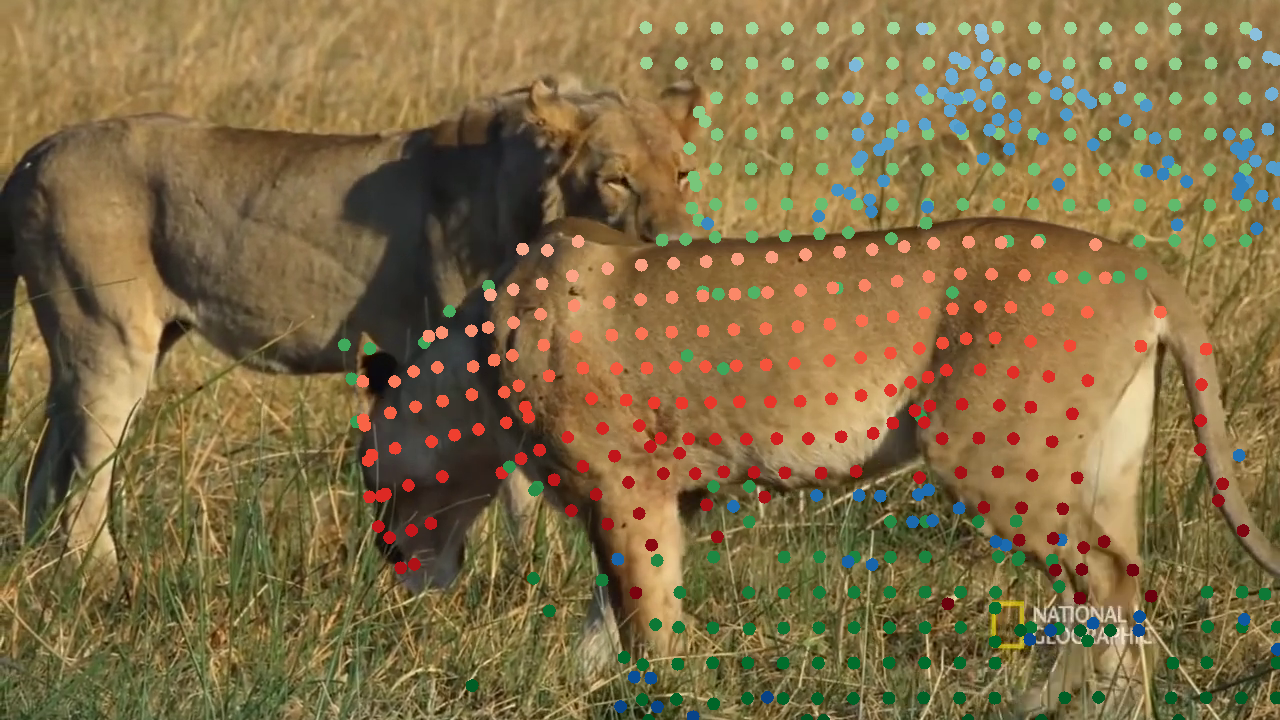}
    \caption{Combined RAFT and RoMa strategy.}
    \label{subfig:ensemble1}
  \end{subfigure}\hfill
  \begin{subfigure}{.45\textwidth}
    \centering
    \includegraphics[width=\linewidth]{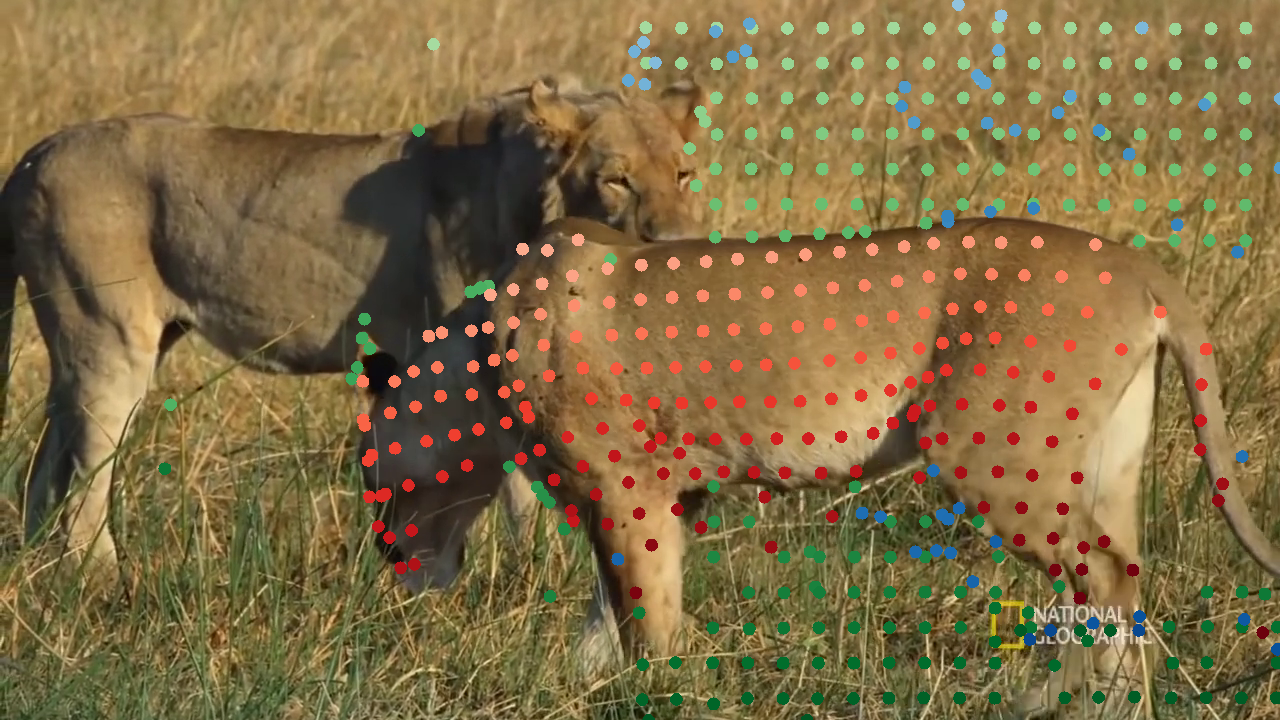}
    \caption{Selective RoMa position prediction.}
    \label{subfig:ensemble2}
  \end{subfigure}

\end{minipage}

\caption{Visual comparison of selected dense tracking methods: (a) reference frame \#0; (b)-(h) predicted positions of points in frame \#140. All blue points are invisible in frame \#140; blue points in (b)-(h) thus indicate false matches. Green points are visible both in frame \#0 and frame \#140. Red points highlight the points on the body of the lioness. Different shades are used to identify different points. The sequence is available at \url{https://cmp.felk.cvut.cz/~serycjon/MFT/visuals/ugsJtsO9w1A-00.00.24.457-00.00.29.462_HD.mp4}.}
\label{fig:raftvsroma}
\end{figure*}

For a visual comparison between the original MFT and the integration of RoMa into MFT, see Figure~\ref{fig:raftvsroma}.

\subsection{Ensembling}\label{subseq:ensembling}

We observed that, in terms of occlusion prediction, MFT's modification of RAFT achieves higher accuracy compared to RoMa. Conversely, RoMa exhibits better performance in optical flow prediction relative to RAFT. Based on these findings, we developed an integrated approach that combines the strengths of both methods. Specifically, our method utilizes occlusion data from RAFT, while RoMa is employed for position prediction, with both processes executed in parallel within the MFT framework. As detailed in Section~\ref{sec:experiments}, our most effective strategy involves employing RAFT for occlusion score prediction and RoMa for position prediction, provided the point is not predicted as occluded; in cases of occlusion, RAFT's predictions are preferred.


\section{Experiments}\label{sec:experiments}

\begin{table*}

\begin{center}
\begin{tabular}{llllllllllll}\toprule
  \multicolumn{2}{c}{}               & \phantom{a} & \multicolumn{3}{c}{main metrics}      & \phantom{a} & \multicolumn{5}{c}{}                                                                        \\
  \cmidrule{4-6} 
  base                 & strategy    &             & AJ          & \deltaavg  & OA         &             & $<\!1$     & $<\!2$     & $<\!4$     & $<\!8$     & $<\!16$    \\
\hline
                       & direct      &             & 38.4        & 50.8       & 65.6       &             & 29.0       & 44.1       & 54.6       & 60.4       & 65.7       \\
RAFT                   & chain       &             & 38.7        & 55.0       & 69.5       &             & 25.2       & 43.8       & 59.4       & 70.4       & 76.3       \\
  \rowcolor{gray!20}
                       & {\bf MFT}   &             & {\bf 47.4}  & {\bf 67.1} & {\bf 77.7} &             & {\bf 34.0} & {\bf 57.3} & {\bf 74.3} & {\bf 82.8} & {\bf 86.9} \\
\hline
                       & chain       &             & 27.3        & 63.5       & 48.2       &             & 36.4       & 56.2       & 69.4       & 76.0       & 79.6       \\
DKM                    & direct      &             & 34.0        & 60.7       & 52.8       &             & 37.0       & 54.5       & 65.3       & 70.9       & 76.0       \\
  \rowcolor{gray!20}
                       & {\bf MFT}   &             & {\bf 47.8}  & {\bf 72.0} & {\bf 70.2} &             & {\bf 43.0} & {\bf 65.8} & {\bf 79.0} & {\bf 84.5} & {\bf 87.8} \\
\hline
                       & direct      &             & 37.7        & 63.7       & 57.6       &             & 37.5       & 55.9       & 67.8       & 75.5       & 81.5       \\
RoMa                   & chain       &             & 40.3        & 63.1       & 60.7       &             & 36.8       & 55.3       & 68.1       & 75.5       & 79.8       \\
  \rowcolor{gray!20}
                       & {\bf MFT}   &             & {\bf 48.8}  & {\bf 72.7} & {\bf 71.7} &             & {\bf 43.0} & {\bf 65.5} & {\bf 79.2} & {\bf 85.5} & {\bf 90.1} \\
  \bottomrule
\end{tabular}
\end{center}
\caption{
\label{tab:MFT_better_than_baselines_first} 
{\bf TAP-Vid DAVIS evaluation of different optical flow combination strategies.}
The MFT strategy outperforms both simple chaining and direct matching for all base optical flow methods on all the metrics.
}
\end{table*}

In this section, we evaluate our proposed method. Initially, we compare the MFT framework with direct optical flow prediction and simple optical flow chaining. Subsequently, we explore RoMa's optical flow prediction performance within the MFT framework depending on whether it predicts the point as occluded or non-occluded, which serves as a foundational finding for our most effective ensembling strategy. The final part of our experimentation serves as a comparison of different ensembling strategies, justifying the design of our most effective architecture, and comparing it to other tracking methods.

\paragraph{Evaluation setup}
Our experiments were conducted on all 30 tracks of the TAP-Vid-DAVIS dataset~\cite{doersch2022tap} with a resolution of 512×512 using the \textit{first} evaluation mode. This approach aligns with the methodology described in MFT~\cite{neoral2024mft}. It is important to stress that in the dataset, the tracks are annotated only sparsely with more focus on the foreground objects rather than the static background.

\paragraph{Evaluation metrics}
In assessing the performance of our approach, we employ three key metrics as defined by the TAP-Vid benchmark. The Occlusion Accuracy (OA) evaluates the accuracy of classifying the points as occluded. We measure the quality of the predicted positions, using average displacement error, denoted as \deltaavg. This metric calculates the fraction of visible points with a positional error below specific thresholds, averaged over thresholds of 1, 2, 4, 8, and 16 pixels. These accuracies for individual thresholds are denoted as $<i$ with $i$ representing the threshold. Additionally, the Average Jaccard (AJ) as defined in \cite{doersch2022tap} index is used to collectively assess both occlusion and position accuracy.

\subsection{MFT Chaining}

A key aspect of our analysis involves contrasting the performance of RAFT, DKM, and RoMa within the MFT framework against \textit{direct} optical flow prediction with the first frame serving as a reference, and \textit{chaining} of the optical flows computed on consecutive video frames. The results presented in Table~\ref{tab:MFT_better_than_baselines_first} clearly show that for each base method (RAFT, DKM, RoMa), the MFT strategy consistently outperforms the other strategies in all metrics by a large margin. These results underscore the effectiveness of MFT in handling complex motion trajectories over extended periods, surpassing the limitations of direct prediction and simple chaining methods. A key observation exemplified in Figure~\ref{fig:raftvsroma} is that RoMa is substantially less prone to predict mismatches in the background than RAFT. 

The results in Table~\ref{tab:MFT_better_than_baselines_first} also show that RoMa within the MFT paradigm achieves arguably the best results in position prediction, while RAFT outperforms all other methods in the occlusion classification accuracy. This finding serves as a foundation for our ensemble strategies in Subsec.~\ref{subseq:ensembling}. Due to the consistently better performance of RoMa over DKM in the evaluation benchmark in all, average Jaccard, average displacement error, and occlusion accuracy we from now on focus our experiments on RoMa even if DKM runs slightly faster.


\subsection{RoMa Visibility}\label{subsec:visibility}

While RoMa demonstrates high accuracy in position prediction, its capability in occlusion detection is relatively limited in comparison to RAFT. However, the quality of occlusion prediction is vital for scoring the optical flows as described in Subsec.~\ref{subsec:chaining}, and thus for computing new flows. We hence conjecture that if we only use the RoMa's optical flow predictions that are predicted as not occluded, we can achieve even better tracking results. The results, as shown in Tab.~\ref{tab:roma_better_on_visible}, indicate a marked improvement in tracking accuracy when measured only on points predicted as non-occluded.

\begin{table}
\begin{center}
\resizebox{\columnwidth}{!}{
\begin{tabular}{lllllll}\toprule
predicted & \deltaavg  & $<\!1$     & $<\!2$     & $<\!4$     & $<\!8$     & $<\!16$    \\
\hline
occluded  & 47.4       & 18.7       & 32.7       & 52.0       & 62.6       & 71.1       \\
  \rowcolor{gray!20}
visible   & {\bf 77.2} & {\bf 46.9} & {\bf 70.9} & {\bf 84.5} & {\bf 89.8} & {\bf 93.7} \\
 any      & 72.7       & 43.0       & 65.5       & 79.2       & 85.5       & 90.1       \\
  \bottomrule
\end{tabular}
}
\end{center}
\caption{
\label{tab:roma_better_on_visible} 
{\bf TAP-Vid DAVIS evaluation of MFT-RoMa separated by the occlusion prediction.}
Using only the points predicted as not occluded leads to improved position accuracy on all error thresholds.
}
\end{table}

\begin{table*}

\begin{center}
\begin{tabular}{llllllllll}\toprule
&\multicolumn{2}{c}{MFT base} & \phantom{a} & \multicolumn{3}{c}{main metrics} & \phantom{a} & \multicolumn{2}{c}{visibility}                       \\
  \cmidrule{2-3} \cmidrule{5-7} \cmidrule{9-10}
&position                     & occlusion   &                                  & AJ          & \deltaavg  & OA         &  & precision  & recall     \\
\hline
(1) & RAFT                         & RAFT        &                                  & 47.4        & 67.1       & {\bf 77.7} &  & {\bf 78.0} & {\bf 91.5} \\
(2) & RoMa                         & RoMa        &                                  & {\bf 48.8}  & {\bf 72.7} & 71.7       &  & 74.5       & 85.3       \\
  \hline
(3) & RoMa                         & RAFT        &                                  & 50.2        & 72.7       & 77.7       &  & 78.0       & 91.5       \\
  \rowcolor{gray!20}
(4) & RAFT/RoMa                    & RAFT        &                                  & {\bf 51.6}  & {\bf 73.4} & {\bf 77.7} &  & {\bf 78.0} & {\bf 91.5} \\
& TAPIR   &&     & 56.2 &  70.0 & 86.5 & & & \\
& CoTracker   &&     & 61.0 &  75.9 & 89.4 & & & \\
  \bottomrule
\end{tabular}
\end{center}

\caption{
\label{tab:ensemble_first} 
{\bf TAP-Vid DAVIS evaluation of combinations of two trackers.}
We run MFT-RAFT and MFT-RoMa independently in parallel, using the two outputs for the final position and occlusion prediction.
RAFT-based MFT (1) has good occlusion accuracy (OA), RoMa-based MFT (2) has good position accuracy \deltaavg.
Using MFT-RAFT to predict occlusion and MFT-RoMa to predict position (3) achieves better AJ.
The best results (4) are achieved when the position is predicted by MFT-RoMa, but only when it predicts visible (see Tab.~\ref{tab:roma_better_on_visible}).
}
\end{table*}

\subsection{Ensembling Strategies}\label{subsec:ensembles}

In the concluding part of our experimental analysis, we compare various ensembling strategies within the MFT framework, building on the insights from the previous sections. The results, detailed in Table~\ref{tab:ensemble_first}, demonstrate the effectiveness of the ensemble strategy.\vspace{-1em}

\paragraph{RAFT-based MFT Strategy}
For comparison we show the original MFT strategy, utilizing RAFT for both position and occlusion predictions. This approach, while achieving the highest occlusion accuracy among all ensembling strategies tested, exhibits suboptimal performance in position precision.\vspace{-1em}

\paragraph{RoMa-based MFT Strategy}
Substituting RAFT entirely with RoMa, we observed an improvement in position prediction accuracy. However, this modification led to a significant decrease in occlusion prediction accuracy, highlighting the trade-offs between these two aspects.\vspace{-1em}

\paragraph{Combined RAFT and RoMa Strategy}
Our next strategy involved a simple combination of RAFT and RoMa: RAFT for occlusion prediction and RoMa for position prediction. This hybrid approach resulted in enhanced performance across all metrics, outperforming the aforementioned individual strategies.\vspace{-1em}

\begin{figure}
\centering

\begin{subfigure}[b]{0.87\columnwidth}
  \centering
  \includegraphics[width=0.45\columnwidth]{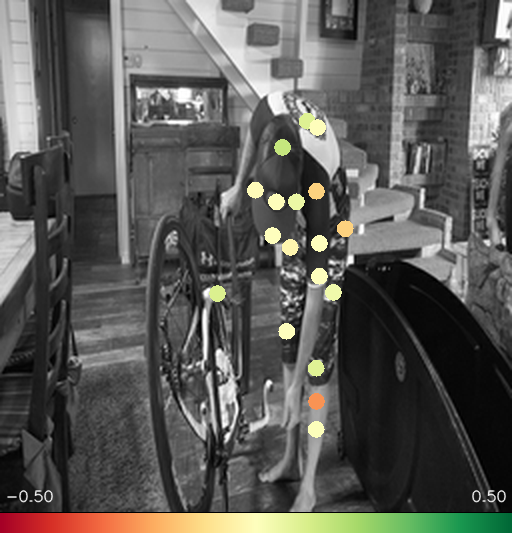}%
  \hfill
  \includegraphics[width=0.45\columnwidth]{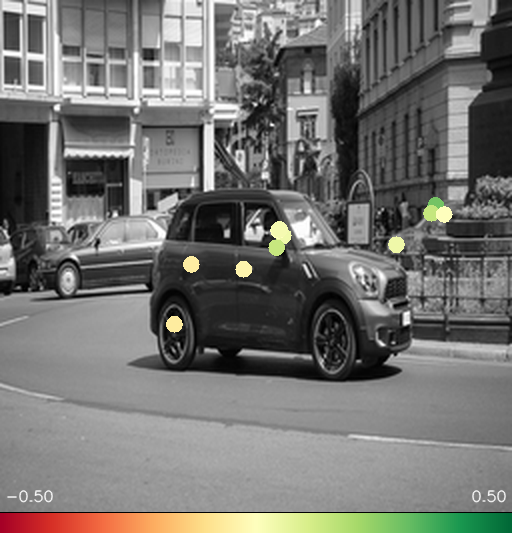}
  \caption{RAFT-based MFT Strategy}
  \label{fig:first-pair}
\end{subfigure}
\hfill 
\begin{subfigure}[b]{0.87\columnwidth}
  \centering
  \includegraphics[width=0.45\columnwidth]{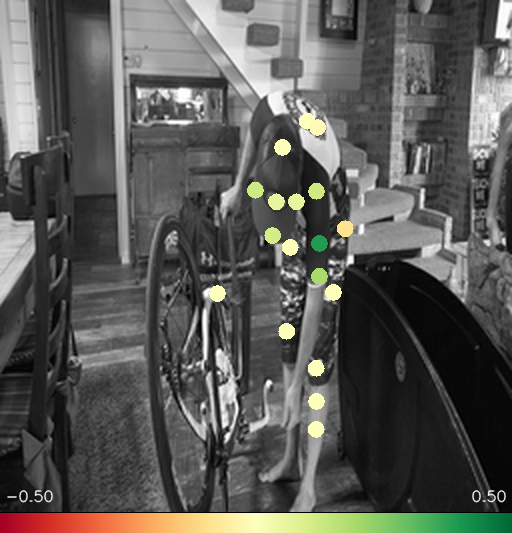}%
  \hfill%
  \includegraphics[width=0.45\columnwidth]{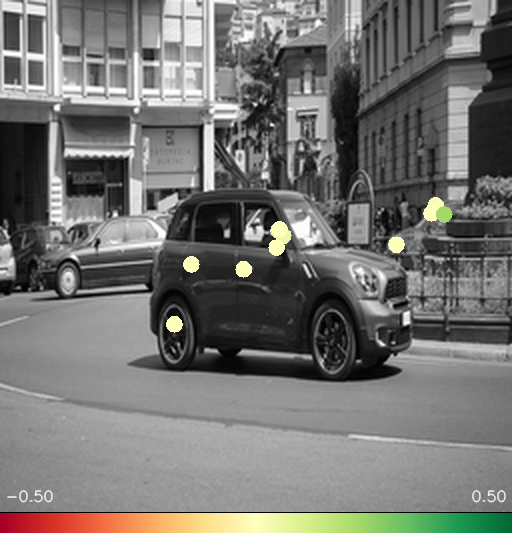}
  \caption{RoMa-based MFT Strategy}
  \label{fig:second-pair}
\end{subfigure}


\begin{subfigure}[b]{0.87\columnwidth}
  \centering
  \includegraphics[width=0.45\columnwidth]{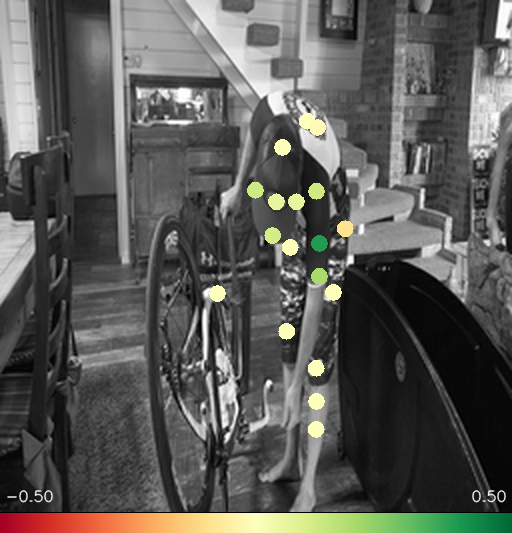}%
  \hfill%
  \includegraphics[width=0.45\columnwidth]{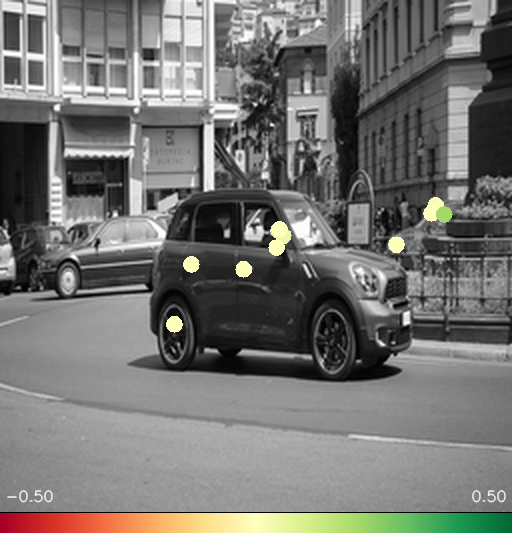}
  \caption{Combined RAFT and RoMa Strategy}
  \label{fig:third-pair}
\end{subfigure}
\caption{Images show the first frames of two selected TAP-Vid DAVIS sequences. Dots represent ground-truth tracking points, with shades of green showing the improvement in \deltaavg \: achieved by the Selective RoMa Position Prediction ensemble over methods (a)-(c), shades of red show the converse.}
\label{fig:tapvidcompare}
\end{figure}

\paragraph{Selective RoMa Position Prediction}
However, further refinement was achieved by integrating findings from Subsection~\ref{subsec:visibility}. We found that RoMa's position predictions are more accurate for points it identifies as visible. Therefore, we devised a strategy where MFT-RoMa's position predictions are used only if the points are marked as visible; otherwise, RAFT's predictions are utilized. This selective strategy led to improvements in both position prediction accuracy and occlusion accuracy. We visually compare this strategy with other two best-performing strategies and MFT with RAFT in Figure \ref{fig:tapvidcompare}.


\paragraph{Comparison with Point Trackers}
We observe that our approach closely rivals or exceeds the performance of established sparse point tracking methods like CoTracker and TAPIR in the average position accuracy while achieving worse performance in the occlusion prediction accuracy. It is noteworthy that our method attains these results within a strictly causal framework, contrasting with CoTracker and TAPIR, which utilize attention-based temporal refinement strategies. Moreover, it is important to highlight that, unlike our approach, CoTracker and TAPIR are designed as sparse trackers.

\section{Conclusion}


We have showcased the benefits of employing the MFT framework over direct optical flow computation and optical flow chaining. We have also demonstrated the flexibility of the MFT paradigm which can be readily used together with different optical flow computation methods. Without complex architectural modifications and using simple ensemble strategies, we were able to demonstrate position prediction accuracy on the Tap-Vid dataset competing with that of state-of-the-art sparse trackers that utilize non-causal tracking refinement.

\paragraph{Limitations and Future Work}\vspace{-0.65em}
Our current approach does not take into account the speed of the baseline optical flow networks. The main limitation is the need for two optical flow networks to operate concurrently within the ensemble strategy. Exploring co-training strategies that enable a single network to deliver similar performance could be a viable solution. A key task is to bridge the existing gap in occlusion prediction accuracy between our method and the state-of-the-art. We also put forward the need for new datasets featuring dense annotations of point tracks in both the foreground and background.

\paragraph{Acknowledgments}
This work was supported by Toyota Motor Europe and
by the Grant Agency of the Czech Technical University in Prague, grant 
No.SGS23/173/OHK3/3T/13.

{\small
\bibliographystyle{ieee}
\bibliography{tomasovo}
}

\end{document}